%% file: main.tex
\documentclass[fleqn]{article} 

\PassOptionsToPackage{numbers, compress}{natbib}

\usepackage[preprint]{neurips_2023}
\usepackage{algorithm}
\usepackage{algpseudocode}

\usepackage[utf8]{inputenc} 
\usepackage[T1]{fontenc}    
\usepackage{hyperref}       
\usepackage{url}            
\usepackage{booktabs}       
\usepackage{amsfonts}       
\usepackage{nicefrac}       
\usepackage{microtype}      
\usepackage{xcolor}         
\usepackage{mathtools}
\usepackage{enumerate}
\input{hayden_preamble}
\newtheorem{remark}{Remark}
\newtheorem{corollary}{Corollary}

\newtheorem{theorem}{Theorem}
\newtheorem{lemma}{Lemma}
\newtheorem{assumption}{Assumption}

\usepackage{hyperref}
\usepackage{url}

\newcounter{TesCounter}

\title{Consistent estimation of generative model representations in the data kernel perspective space}

\author{
Aranyak Acharyya
\\
Johns Hopkins University \\
\texttt{aachary6@jhu.edu}
\And
Michael~W.~Trosset \\
Indiana University \\
\texttt{mtrosset@indiana.edu}
\And
Carey~E.~Priebe \\
Johns Hopkins University \\
\texttt{cep@jhu.edu}
\And
Hayden~S.~Helm\\
Helivan Research \\
\texttt{hayden@helivan.io}
}

\begin{document}
\maketitle

\begin{abstract}  
    Generative models, such as large language models and text-to-image diffusion models, produce relevant information when presented a query. 
    Different models may produce different information when presented the same query.
    As the landscape of generative models evolves, it is important to develop techniques to study and analyze differences in model behaviour. 
    In this paper we present novel theoretical results for embedding-based representations of generative models in the context of a set of queries. 
    In particular, we establish sufficient conditions for the consistent estimation of the model embeddings in situations where the query set and the number of models grow.
\end{abstract}

\input{text/introduction}
\input{text/prelims_aistats}

\input{text/setting}

\input{text/consistency_aistats}

\input{text/numerical_aistats}
\input{text/discussion_aitstats}

\subsubsection*{Acknowledgements.} We would like to thank Avanti Athreya, Brandon Duderstadt, Youngser Park, and Zekun Wang for helpful discussions and comments throughout the development of this manuscript.

\clearpage

\bibliographystyle{iclr2021_conference}
\bibliography{ref}

\clearpage

\appendix
\section{Proofs of lemmas and theorems}
\label{Sec:Appendix}
\subsection{Fixed set of models and fixed set of queries}
\textbf{Theorem 1.}
\textit{
Let $\widehat{\boldsymbol{\psi}} \in
    \mathrm{MDS}(\mathbf{D}) \subset \mathbb{R}^{n \times d}$. Then
    there exists a subsequence $\lbrace r_u \rbrace_{u=1}^{\infty}$ of $\lbrace r \rbrace_{r=1}^{\infty}$ such that
    for all pairs $(i,i') \in [n]^2$,
    \begin{equation*}
    \left(
      \left\lVert 
\widehat{\boldsymbol{\psi}}_i^{(r_u)}-
\widehat{\boldsymbol{\psi}}_{i'}^{(r_u)}
      \right\rVert
      -
      \left\lVert 
\boldsymbol{\psi}_i-
\boldsymbol{\psi}_{i'}
      \right\rVert
      \right) \to^P 0
      \text{ as $u \to \infty$},
    \end{equation*}
     where $\boldsymbol{\psi}=
     [\boldsymbol{\psi}_1|\boldsymbol{\psi}_2|\dots | \boldsymbol{\psi}_n]^T
     \in
    \mathrm{MDS}(\boldsymbol{\Delta}) \subset \mathbb{R}^{n \times d}
    $.
}
    \newline
    \newline
    \textbf{Proof.}
    See Theorem 2 of \cite{trosset2024continuous}.
\subsection{Fixed set of models and growing set of queries}
\textbf{Lemma 1.}
\textit{Let $ \widehat{\boldsymbol{\psi}} \in
\mathrm{MDS}(\mathbf{D}) \subset \mathbb{R}^{n \times d}$.
Let $ n $ be fixed and let $ m $ grow with $ r $.
If $\left\lVert \mathbf{D}-\boldsymbol{\Delta}^{(\infty)} \right\rVert_F \to 0$ as $r \to \infty$, then there exists a subsequence $\lbrace r_u \rbrace_{u=1}^{\infty}$ of $\lbrace r \rbrace_{r=1}^{\infty}$ such that
for all pairs $(i,i') \in [n]^2$,
\begin{equation*}
\left(
    \left\lVert 
\widehat{\boldsymbol{\psi}}_i^{(r_u)}-
\widehat{\boldsymbol{\psi}}_{i'}^{(r_u)}
\right\rVert -
\left\lVert 
\boldsymbol{\psi}_i -
\boldsymbol{\psi}_{i'}
\right\rVert
\right) \to 0
\text{ as $u \to \infty$},
\end{equation*}
 where $\boldsymbol{\psi}=
 [\boldsymbol{\psi}_1|\boldsymbol{\psi}_2|\dots |\boldsymbol{\psi}_n]^T
 \in \mathrm{MDS}(\boldsymbol{\Delta}^{(\infty)}) \subset \mathbb{R}^{n \times d}$.
}

\textbf{Proof.} See \textit{Theorem 2} of \cite{trosset2024continuous}.
\newline
\newline
\newline
\textbf{Theorem 2.}
\textit{Let $ \boldsymbol{\Sigma}_{ij}=\mathrm{cov}[g(f_i(q_j)_k)]$ be the covariance matrix associated with the distribution $F_{ij}$ and $\gamma_{ij}=\mathrm{trace}(\boldsymbol{\Sigma}_{ij})$. 
If, for all $i \in [n]$,
\begin{equation*}
   \lim_{r \to \infty}
    \frac{ \frac{1}{m}
    \sum_{j=1}^{m} \gamma_{ij}}{r}=0
\end{equation*}
then
     $
     \left\lVert
\mathbf{D}-\boldsymbol{\Delta}^{(\infty)}  
\right\rVert_F \to^P 0
     $ as $r \to \infty$.
}
\newline
\textbf{Proof.} Observe that by triangle inequality,
\begin{equation*}
    \begin{aligned}
        &\left\lVert 
\bar{\mathbf{X}}_{i}-\bar{\mathbf{X}}_{i'}
        \right\rVert_F \leq 
        \left\lVert 
(\bar{\mathbf{X}}_{i}-\boldsymbol{\mu}_{i})-
(
\bar{\mathbf{X}}_{i'}-\boldsymbol{\mu}_{i'})
        \right\rVert_F +
        \left\lVert 
\boldsymbol{\mu}_{i}-\boldsymbol{\mu}_{i'}
        \right\rVert_F \\
&\implies 
\left\lVert 
\bar{\mathbf{X}}_{i}-\bar{\mathbf{X}}_{i'}
        \right\rVert_F -
\left\lVert 
\boldsymbol{\mu}_{i}-\boldsymbol{\mu}_{i'}
        \right\rVert_F   \leq 
\left\lVert 
\bar{\mathbf{X}}_{i}-\boldsymbol{\mu}_{i}
\right\rVert_F
+
\left\lVert
\bar{\mathbf{X}}_{i'}-\boldsymbol{\mu}_{i'}
        \right\rVert_F.
    \end{aligned}
\end{equation*}
Similarly, we obtain,
\begin{equation*}
    \begin{aligned} 
        &\left\lVert 
\boldsymbol{\mu}_{i'}-\boldsymbol{\mu}_{i}
        \right\rVert_F \leq 
        \left\lVert 
(\bar{\mathbf{X}}_{i}-\boldsymbol{\mu}_{i})-
(\bar{\mathbf{X}}_{i'}-\boldsymbol{\mu}_{i'})
        \right\rVert_F +
        \left\lVert 
\bar{\mathbf{X}}_{i'}-\bar{\mathbf{X}}_{i}
        \right\rVert_F \\
&\implies 
\left\lVert 
\boldsymbol{\mu}_{i'}-\boldsymbol{\mu}_{i}
        \right\rVert_F -
\left\lVert 
\bar{\mathbf{X}}_{i'}-\bar{\mathbf{X}}_{i}
        \right\rVert_F   \leq 
\left\lVert 
\bar{\mathbf{X}}_{i}-\boldsymbol{\mu}_{i}
\right\rVert_F
+
\left\lVert
\bar{\mathbf{X}}_{i'}-\boldsymbol{\mu}_{i'}
        \right\rVert_F.
    \end{aligned}
\end{equation*}
Thus, from the above two equations combined, we get,
\begin{equation}
    \left|
    \left\lVert 
\bar{\mathbf{X}}_{i}-\bar{\mathbf{X}}_{i'}
        \right\rVert_F -
\left\lVert 
\boldsymbol{\mu}_{i}-\boldsymbol{\mu}_{i'}
        \right\rVert_F 
    \right| \leq 
\left\lVert 
\bar{\mathbf{X}}_{i}-\boldsymbol{\mu}_{i}
\right\rVert_F
+
\left\lVert
\bar{\mathbf{X}}_{i'}-\boldsymbol{\mu}_{i'}
        \right\rVert_F.
\end{equation}
\newline
From eqn $(1)$,
\begin{equation}
\left|
\mathbf{D}_{i i'}-
\boldsymbol{\Delta}_{i i'}
\right| 
\leq 
\frac{1}{m}
\left\lVert 
\bar{\mathbf{X}}_i-
\boldsymbol{\mu}_i
\right\rVert_F +
\frac{1}{m}
\left\lVert 
\bar{\mathbf{X}}_{i'}-
\boldsymbol{\mu}_{i'}
\right\rVert_F.
\end{equation}
Eqn $(2)$ tells us that, in order
to get $\left\lVert \mathbf{D}-\boldsymbol{\Delta} \right\rVert_F \to^P 0$, it suffices to have 
$
\frac{1}{m} 
\left\lVert 
\bar{\mathbf{X}}_{i}-
\boldsymbol{\mu}_i
\right\rVert_F \to^P 0
$ for all $i \in [n]$.
Now, for an arbitrary $\epsilon$,
\begin{equation*}
\begin{aligned}
    &\mathbb{P}\left[
    \frac{1}{m}
    \left\lVert 
\bar{\mathbf{X}}_i -
\boldsymbol{\mu}_i
    \right\rVert_F > \epsilon
    \right] \\ &\leq 
    \mathbb{P}
    \left[
\sum_{j=1}^m 
\left\lVert
(\bar{\mathbf{X}}_i)_{j \cdot}
-
(\boldsymbol{\mu}_i)_{j \cdot}
\right\rVert^2 > m^2 \epsilon^2
    \right] \\ &\leq 
\mathbb{P} 
\left[
\bigcup_{j=1}^m 
\left\lbrace 
\left\lVert 
(\bar{\mathbf{X}}_i)_{j \cdot}
-
(\boldsymbol{\mu}_i)_{j \cdot}
\right\rVert^2 > m \epsilon^2
\right\rbrace
\right] \\
&\leq 
\sum_{j=1}^m 
\mathbb{P}
\left[
\left\lVert 
(\bar{\mathbf{X}}_i)_{j \cdot}
-
(\boldsymbol{\mu}_i)_{j \cdot}
\right\rVert^2 > m \epsilon^2
\right] \\
&\leq
\sum_{j=1}^m
\frac{
\mathbb{E}
\left\lVert 
(\bar{\mathbf{X}}_i)_{j \cdot}-
(\boldsymbol{\mu}_i)_{j \cdot}
\right\rVert^2
}
{
m \epsilon^2
}.
\end{aligned}
\end{equation*}
Recall that $(\bar{\mathbf{X}}_i)_{j \cdot}=
\frac
{1}{r} \sum_{k=1}^r \mathbf{X}_{ijk}
$ where $\mathbf{X}_{ijk} \sim^{iid} F_{ij}$ and the distribution $F_{ij}$ has mean $(\boldsymbol{\mu}_i)_{j \cdot}$ and covariance matrix $\boldsymbol{\Sigma}_{ij}$. Thus, 
$
    \mathbb{E}[(\bar{\mathbf{X}}_i)_{j \cdot}]=(\boldsymbol{\mu}_i)_{j \cdot}$
and
$
    \mathrm{cov}[(\bar{\mathbf{X}}_i)_{j \cdot}]=\frac{1}{r}\boldsymbol{\Sigma}_{ij}
$.
Hence,
\begin{equation*}
    \mathbb{E}
    \left\lVert 
(\bar{\mathbf{X}}_i)_{j \cdot}-
(\boldsymbol{\mu}_i)_{j \cdot}
    \right\rVert^2
    =
    \mathrm{trace}(\mathrm{cov}[(\bar{\mathbf{X}}_i)_{j \cdot}])
    =\mathrm{trace}\left(\frac{1}{r} \boldsymbol{\Sigma}_{ij} \right)=\frac{\gamma_{ij}}{r}
\end{equation*}
where $\gamma_{ij}=\mathrm{trace}(\boldsymbol{\Sigma}_{ij})$.
 Hence, 
\begin{equation*}
    \mathbb{P}\left[
    \frac{1}{m}
    \left\lVert 
\bar{\mathbf{X}}_i -
\boldsymbol{\mu}_i
    \right\rVert_F > \epsilon
    \right] \leq 
\sum_{j=1}^m 
\frac{
\mathbb{E}
\left\lVert 
(\bar{\mathbf{X}}_i)_{j \cdot}-
(\boldsymbol{\mu}_i)_{j \cdot}
\right\rVert^2
}
{
m \epsilon^2
} 
\leq 
\frac{
\sum_{j=1}^m \gamma_{ij}
}{r m \epsilon^2}.
\end{equation*}
Recall that as 
$r \to \infty$,
$\left\lVert \mathbf{D}-\boldsymbol{\Delta} \right\rVert_F \to^P 0$ if 
$
\frac{1}{m}
\left\lVert 
\bar{\mathbf{X}}_i-\boldsymbol{\mu}_i
\right\rVert_F \to^P 0
$ for all $i \in [n]$.
\newline
Thus, 
$\lim_{r \to \infty}
\frac{
\frac{1}{m} \sum_{j=1}^m \gamma_{ij}
}{r}
=
0
$ for all $i \in [n]$ ensures that 
$\left\lVert \mathbf{D}-\boldsymbol{\Delta} \right\rVert \to^P 0$, which again implies (using \textit{Assumption \ref{Asm:Asymptotic_Euclidean_realizability_fixed_n}} and Triangle Inequality)
$\left\lVert 
\mathbf{D}-\boldsymbol{\Delta}^{(\infty)}
\right\rVert \to^P 0$ as $r \to \infty$.
\newline
\newline
\newline
\textbf{Theorem 3.}
\textit{
 In the setting of \textit{Theorem \ref{Th:sufficient_condition}},
    suppose for all $i \in [n]$, $\frac{1}{m} \sum_{j=1}^{m} \gamma_{ij}=
o(r)$.
Also, let $\widehat{\boldsymbol{\psi}} \in 
\mathrm{MDS}(\mathbf{D}) \subset \mathbb{R}^{n \times d}$.
There exists a subsequence $\lbrace r_u \rbrace_{u=1}^{\infty}$ of $\lbrace r \rbrace_{r=1}^{\infty}$ such that
 for all $(i,i') \in [n]^2$, 
 \begin{equation*}
 \left(
     \left\lVert 
\widehat{\boldsymbol{\psi}}_i^{(r_u)}-
\widehat{\boldsymbol{\psi}}_{i'}^{(r_u)}
\right\rVert -
\left\lVert 
\boldsymbol{\psi}_i
-
\boldsymbol{\psi}_{i'}
\right\rVert
\right) \to^P 0
\text{ as $u \to \infty$},
 \end{equation*}
  where $\boldsymbol{\psi}=
  [\boldsymbol{\psi}_1|\boldsymbol{\psi}_2|\dots |\boldsymbol{\psi}_n]^T
  \in \mathrm{MDS}(\boldsymbol{\Delta}^{(\infty)}) \subset \mathbb{R}^{n \times d}$.
}
\newline
\textbf{Proof.} 
Let us assume
$\frac{1}{m}\sum_{j=1}^m \gamma_{ij}=o(r)$.
From \textit{Theorem 2},
then we must have
$
\left\lVert 
\mathbf{D}-
\boldsymbol{\Delta}^{(\infty)}
\right\rVert_F \to^P 0
$ as $r \to \infty$. From \textit{Lemma 1}, this means 
there exists a subsequence $\lbrace
r_u
\rbrace_{u=1}^{\infty}$ of $\lbrace r \rbrace_{r=1}^{\infty}$ such that for all pairs $(i,i') \in [n]^2$, 
\begin{equation*}
\left(
    \left\lVert 
\widehat{\boldsymbol{\psi}}_i^{(r_u)}-
\widehat{\boldsymbol{\psi}}_{i'}^{(r_u)}
    \right\rVert -
    \left\lVert 
\boldsymbol{\psi}_i-
\boldsymbol{\psi}_{i'}
    \right\rVert
    \right) \to^P 0
\end{equation*}
as $u \to \infty$.

\subsection{Growing set of models and growing set of queries}
\textbf{Lemma 2.}
\textit{
Let $\boldsymbol{\phi}_i \sim^{iid} \mathcal{P}$. Assume that for all pairs $(i,i') \in \mathbb{N} \times \mathbb{N}$, 
    $\mathbf{D}_{i i'} \to^P \boldsymbol{\Delta}^{(\infty)}(\boldsymbol{\phi}_i,\boldsymbol{\phi}_{i'})$ as $r \to \infty$.
    Then, for some subsequence $\lbrace r_u \rbrace_{u=1}^{\infty}$ of $\lbrace r \rbrace_{r=1}^{\infty}$, for all $p \geq 1$,
\begin{equation*}
\int_{\mathcal{M}}
\int_{\mathcal{M}}
\left(
\left\lVert 
\widehat{\boldsymbol{\psi}}_1^{(r_u)}-
\widehat{\boldsymbol{\psi}}_2^{(r_u)}
\right\rVert -
\left\lVert 
\mathrm{mds}(\boldsymbol{\phi}_1)-
\mathrm{mds}(\boldsymbol{\phi}_2)
\right\rVert
\right)^p
\mathcal{P}(d \boldsymbol{\phi}_1)
\mathcal{P}(d \boldsymbol{\phi}_2)
\to^P
0 \text{ as $u \to \infty$}.
\end{equation*}
}
\newline
\newline
\textbf{Proof.} See \textit{Theorem 3} of \cite{trosset2024continuous}.
\newline
\newline
\newline
\textbf{Theorem 4.}
In our setting where $m,n \to \infty$ as $r \to \infty$, $|\mathbf{D}_{i i'}-\boldsymbol{\Delta}^{(\infty)}
(\boldsymbol{\phi}_i,
\boldsymbol{\phi}_{i'})
| \to^P 0$ for every pair $(i,i') \in \mathbb{N} \times \mathbb{N}$, if 
\begin{equation*}
    \lim_{r \to \infty}
    \frac
    {
    \frac{1}{m} \sum_{j=1}^m \gamma_{ij}
    }
    {
    r
    } =0
    \hspace{0.2cm} 
    \text{for all $i$}.
\end{equation*}
\newline
\textbf{Proof.}
Recall (from the proof of \textit{Theorem 2}) that 
$
\left\lvert 
\mathbf{D}_{i i'}-
\boldsymbol{\Delta}_{i i'}
\right\rvert
\leq 
\frac{1}{m}
\left\lVert 
\bar{\mathbf{X}}_i-
\boldsymbol{\mu}_i
\right\rVert_F +
\frac{1}{m}
\left\lVert 
\bar{\mathbf{X}}_{i'}-
\boldsymbol{\mu}_{i'}
\right\rVert_F
$
and
 that for all $i$, 
$
\mathbb{P}
\left[
\frac{1}{m}
\left\lVert 
\bar{\mathbf{X}}_i-
\boldsymbol{\mu}_i
\right\rVert_F
>\epsilon
\right]
\leq 
\frac{
\sum_{j=1}^m \gamma_{ij}
}{r m \epsilon^2}
$ for any $\epsilon>0$. Thus, 
if 
$
\lim_{r \to \infty}
\frac
{
\frac{1}{m}\sum_{j=1}^m \gamma_{ij}
}{r}=0
$ for all $i$, then 
$\frac{1}{m}
\left\lVert 
\bar{\mathbf{X}}_i-
\boldsymbol{\mu}_i
\right\rVert \to^P 0
$ for all $i$, which in turn implies 
$
|\mathbf{D}_{i i'}-
\boldsymbol{\Delta}_{i i'}| \to ^P 0
$ for all pairs $(i,i') \in \mathbb{N}^2$ as $r \to \infty$. Using \textit{Assumption \ref{Asm:Asymptotic_Euclidean_realizability_growing_n}} and Triangle Inequality, we have that for all pairs $(i,i') \in \mathbb{N}^2$, 
$|\mathbf{D}_{i i'}-
\boldsymbol{\Delta}^{(\infty)}(\boldsymbol{\phi}_i,\boldsymbol{\phi}_{i'})| \to^P 0$ as $r \to \infty$. Using \textit{Lemma 2}, for all $p \geq 1$, for some subsequence 
$\lbrace r_u \rbrace_{u=1}^{\infty}$ of $\lbrace r \rbrace_{r=1}^{\infty}$, 
\begin{equation*}
\int_{\mathcal{M}}
\int_{\mathcal{M}}
\left(
\left\lVert 
\widehat{\boldsymbol{\psi}}_1^{(r_u)}-
\widehat{\boldsymbol{\psi}}_2^{(r_u)}
\right\rVert -
\left\lVert 
\mathrm{mds}(\boldsymbol{\phi}_1)-
\mathrm{mds}(\boldsymbol{\phi}_2)
\right\rVert
\right)^p
\mathcal{P}(d \boldsymbol{\phi}_1)
\mathcal{P}(d \boldsymbol{\phi}_2)
\to^P
0 \text{ as $u \to \infty$}.
\end{equation*}
\newline
\newline
\newline
\textbf{Theorem 5.}
\textit{
In the setting of \textit{Lemma \ref{Lm:D_to_psi_consistency_growing_n}}, 
    suppose for all $i \in \mathbb{N}$, 
    $\frac{1}{m} \sum_{j=1}^m \gamma_{ij}=o(r)$. Then,
    for all $p \geq 1$, for some subsequence $\lbrace r_u \rbrace_{u=1}^{\infty}$ of $\lbrace r \rbrace_{r=1}^{\infty}$,
    \begin{equation*}
\int_{\mathcal{M}}
\int_{\mathcal{M}}
        \left(
        \left\lVert 
\widehat{\boldsymbol{\psi}}_1^{(r_u)}-
\widehat{\boldsymbol{\psi}}_2^{(r_u)}
        \right\rVert -
        \left\lVert 
\mathrm{mds}(\boldsymbol{\phi}_1) -
\mathrm{mds}(\boldsymbol{\phi}_2)
        \right\rVert
        \right)^p
        \mathcal{P}(d \boldsymbol{\phi}_1)
        \mathcal{P} (d \boldsymbol{\phi}_2)
        \to^P
        0
        \text{ as $u \to \infty$}.
    \end{equation*}
}
    \newline
    \textbf{Proof.} 
    Let us assume 
    that 
    $
    \frac{1}{m}
    \sum_{j=1}^m \gamma_{ij}=o(r)
    $. Then, by \textit{Theorem \ref{Th:sufficient_condition_growing_n}},  for all pairs $(i,i') \in \mathbb{N}^2$, 
    $
    |\mathbf{D}_{i i'} -
    \boldsymbol{\Delta}^{(\infty)}(\boldsymbol{\phi}_i,\boldsymbol{\phi}_{i'})| \to^P 0
    $ as $r \to \infty$. Using \textit{Lemma \ref{Lm:D_to_psi_consistency_growing_n}}, for some subsequence $\lbrace r_u \rbrace_{u=1}^{\infty}$ of $\lbrace r \rbrace_{r=1}^{\infty}$, for all $p \geq 1$,
    \begin{equation*}
\int_{\mathcal{M}}
\int_{\mathcal{M}}
\left(
\left\lVert 
\widehat{\boldsymbol{\psi}}_1^{(r_u)}-
\widehat{\boldsymbol{\psi}}_2^{(r_u)}
\right\rVert -
\left\lVert 
\mathrm{mds}(\boldsymbol{\phi}_1)-
\mathrm{mds}(\boldsymbol{\phi}_2)
\right\rVert
\right)^p
\mathcal{P}(d \boldsymbol{\phi}_1)
\mathcal{P}(d \boldsymbol{\phi}_2)
\to^P
0 \text{ as $u \to \infty$}.
\end{equation*}.
\clearpage

\end{document}

%% file: hayden_preamble.tex
\usepackage{caption}
\usepackage{amsthm}

\usepackage{subcaption}
\usepackage{graphicx}

\usepackage{natbib}

\usepackage{amsfonts,amsmath,amssymb}
\usepackage{bbm}
\usepackage{xspace}

\usepackage{multicol}
\usepackage{enumitem}

%% file: text/introduction.tex
\section{Introduction}
\label{Sec:Intro}

Generative models have gained popularity in natural language processing \citep{brown2020language, sanh2021multitask}, text-to-image generation \citep{crowson2022vqgan}, and code generation \citep{zhang2023planning}, as well as in other domains \citep{arik2017deep, singer2022make}. 
The key feature of these models is their ability to generate relevant responses to wide-ranging queries.
Based on differences such as pre-training data mixture, architecture, size, etc., the responses of different models (or of different generation configurations of the same model) may vary widely.
Recently, empirical investigations have demonstrated the potential of embedding-based vector representations of models for capturing meaningful differences in model behavior in the context of a set of queries \citep{faggioli2023perspectives, duderstadt2024comparingfoundationmodelsusing,helm2024tracking}. 
Following \cite{helm2024tracking}, the resulting representations are referred to as \textit{perspectives} of the models and the space in which the representations reside is referred to as the \textit{perspective space} of the models. 


In this paper we provide theoretical justification for these successful empirical investigations by arguing that the estimated perspective space is consistent for a population-level perspective space.
In particular, we analyze three different -- progressively more complicated -- settings in which the perspective space can be used: i) fixed collection of models and fixed set of queries; ii) fixed collection of models and growing set of queries; and iii) growing collection of models and growing set of queries.
For each setting we show that the multi-dimensional scaling with the raw stress criterion of a collection of matrix representations yields vectors that are consistent for an appropriately defined limiting configuration.
Importantly, we provide sufficient conditions for consistency as a function of the generative properties of each model.


Our work is a part of two major bodies of literature: embeddings of inputs and outputs of generative models \citep{mikolov2013efficient, reimers2019sentence, neelakantan2022text, 10098736} and embedding complex objects in suitable Euclidean spaces via multi-dimensional scaling \citep{borg2005modern,WANG2020117274, helm2021inducing,chen2022mental}.
In particular, the perspective space is a minimizer of the raw stress criterion \citep{kruskal1964multidimensional} for a dissimilarity matrix or function \citep{trosset2024continuous} defined on a set of matrix representations of generative models whose rows correspond to the average embedded response for a particular query.






The rest of the paper is organized as follows. 
In \textit{Section} \ref{Sec:Preliminiaries} we introduce notation and review multi-dimensional scaling by the raw stress embedding criterion.
We describe our setting in \textit{Section} \ref{Sec:Description_of_setting} and then establish our main theoretical results in \textit{Section} \ref{Sec:Consistency_of_model_embeddings}.
We provide numerical evidence to support the theoretical results in \textit{Section} \ref{Sec:Numerical_exp} and discuss our results, potential applications, and extensions in \textit{Section} \ref{Sec:Discussion}.

\textbf{Contribution.} We build upon recent theoretical results related to multi-dimensional scaling via the raw stress criterion. 
In particular, we provide sufficient conditions for the consistency of sampling-based vector representations of black-box generative models. 
Our results are general to collections of generative models whose outputs can be mapped to a shared embedding space. 


%% file: text/prelims_aistats.tex
\section{Preliminaries}
\label{Sec:Preliminiaries}

\subsection{Notations}
\label{Subsec:Notations}

Bold letters (such as $\mathbf{B}$ or $\boldsymbol{\mu}$) are used to represent vectors and matrices. Any vector by default is a column vector. For a matrix $\mathbf{B}$, the $j$-th row is denoted by $(\mathbf{B})_{j \cdot}$, and the $(i,i')$-th entry is denoted by $\mathbf{B}_{i i'}$. Moreover, $\left\lVert \mathbf{B} \right\rVert_F$ denotes the Frobenius norm of the matrix $\mathbf{B}$. For any two vectors $\mathbf{x}$ and $\mathbf{y}$, $\left\lVert 
\mathbf{x}-\mathbf{y}
\right\rVert$ denotes the Euclidean distance between $\mathbf{x}$ and $\mathbf{y}$. The set of $d \times d$ orthogonal matrices is denoted by $\mathcal{O}(d)$.

\subsection{Multidimensional scaling by raw stress embedding}
\label{Subsec:Raw_Stress}

\textit{Multidimensional scaling} \citep{borg2005modern} refers to the family of methods that produce vector representations of a set of objects from their pairwise dissimilarities.
Raw stress embedding \citep{kruskal1964multidimensional,trosset2024continuous} is a popular method for multidimensional scaling. 
For our purposes, we discuss raw stress embedding in the context of two scenarios: finite sample size and infinite sample size. 
The following is adapted from \cite{trosset2024continuous} for our specific setting of collections of generative models.          

\textbf{Finite sample size}: 
Let us assume there are $n$ objects and that the matrix
$\boldsymbol{\Delta}^{(\infty)} \in \mathbb{R}^{n \times n}$ is such that $\boldsymbol{\Delta}^{(\infty)}_{ij}$ is the dissimilarity between the $i$-th object and the $j$-th object for $ i, j \in [n] $. 
For a given embedding dimension $d \in \mathbb{N}$, our goal is to find vectors $\boldsymbol{\psi}_1,\dots \boldsymbol{\psi}_n \in \mathbb{R}^d$ such that 
$\left\lVert 
\boldsymbol{\psi}_i-\boldsymbol{\psi}_j
\right\rVert
\approx \boldsymbol{\Delta}^{(\infty)}_{ij}
$. We define solutions to this problem as:
\begin{equation}
\label{Eq:raw_stress_fixed_n}(\boldsymbol{\psi}_1,\dots \boldsymbol{\psi}_n)=
    \arg \min_{\mathbf{z}_i \in \mathbb{R}^d} 
    \sum_{i,i'=1}^n
    \left(
\left\lVert 
\mathbf{z}_i-
\mathbf{z}_{i'}
\right\rVert -
\boldsymbol{\Delta}^{(\infty)}_{i i'}
    \right)^2.
\end{equation}
We let $\boldsymbol{\psi} \in \mathbb{R}^{n \times d}$ denote the matrix whose $i$-th row is $\boldsymbol{\psi}_i$ and write 
$\boldsymbol{\psi}=\mathrm{mds}(\boldsymbol{\Delta}^{(\infty)})$.
\begin{remark}
    Given dissimilarity matrix $\boldsymbol{\Delta}^{(\infty)}$, $\boldsymbol{\psi}=\mathrm{mds}(\boldsymbol{\Delta}^{(\infty)})$ is not  unique -- an affine transformation upon a minimizer gives another minimizer. We denote the set of all solutions $\mathrm{mds}(\boldsymbol{\Delta}^{(\infty)})$ by $\mathrm{MDS}(\boldsymbol{\Delta}^{(\infty)})$. Apart from affine transformations of a solution, there may be two different minimizers that are not affine transformation of each other.
\end{remark}

\begin{remark}
Note that 
\begin{align*}
    \min_{\mathbf{z}_i \in \mathbb{R}^d}    \sum_{i,i'=1}^n
    \left(
\left\lVert 
\mathbf{z}_i-
\mathbf{z}_{i'}
\right\rVert -
\boldsymbol{\Delta}_{i i'}^{(\infty)}
    \right)^2=
    \min_{\tilde{\boldsymbol{\Delta}}}
    \left\lVert
\tilde{\boldsymbol{\Delta}}-\boldsymbol{\Delta}^{(\infty)}
    \right\rVert_F^2,
\end{align*} where 
$\tilde{\boldsymbol{\Delta}}$ is varied over the set of all $n \times n$ Euclidean distance matrices. From the proof of \textit{Theorem 2} in \cite{trosset2024continuous}, it can be argued that it is enough to vary $\tilde{\boldsymbol{\Delta}}$ over 
    $\mathcal{Y}_n$, the set of all $ n \times n $ Euclidean distance matrices
with Frobenius norm less than or equal to $\left\lVert 
\boldsymbol{\Delta}^{(\infty)}
    \right\rVert_F$, which is closed, bounded, and complete. This guarantees the existence of a solution to Eq.\ \eqref{Eq:raw_stress_fixed_n}.
\end{remark}

\textbf{Infinite sample size}: 
In this case, the goal is to find Euclidean representations of all objects from a compact set. 
Since the number of objects is not necessarily finite, we may not be able to arrange the pairwise dissimilarities in a matrix and, instead, rely on the notion of a dissimilarity function.
Let $\mathcal{M}$ be a compact metric space,  $\boldsymbol{\Delta}^{(\infty)}:
\mathcal{M} \times 
\mathcal{M} \to
\mathbb{R}_{\ge 0}
$ be a dissimilarity function on $ \mathcal{M} $, and 
$h:\mathcal{M} \to \mathbb{R}^d$ be a
Borel-measurable
embedding function. 
The \textit{continuous raw stress criterion} \citep{trosset2024continuous} is defined by 
\begin{align}
\label{Eq:continuous_raw_stress}
\nonumber &\sigma((\boldsymbol{\Delta}^{(\infty)},\mathcal{P}),h)= 
\int_{\mathcal{M}}
\int_{\mathcal{M}}
\left(
\left\lVert 
h(\mathbf{m}')-
h(\mathbf{m}'')
\right\rVert -
\boldsymbol{\Delta}^{(\infty)}(\mathbf{m}',\mathbf{m}'')
\right)^2 
  \mathcal{P}(d \mathbf{m}')
\mathcal{P} (d \mathbf{m}'').
\end{align}
We let $\mathrm{mds}:\mathcal{M} \to \mathbb{R}^d$ denote the embedding function that minimizes $\sigma((\boldsymbol{\Delta}^{(\infty)},\mathcal{P}),h)$ by varying $h$ over the set of all possible Borel-measurable embedding functions from $\mathcal{M}$ to $\mathbb{R}^d$.

\begin{remark}
    The function $\tilde{\boldsymbol{\Delta}}: \mathcal{M} \times \mathcal{M} \to \mathbb{R}$ defined as $\tilde{\boldsymbol{\Delta}}(\mathbf{m'},\mathbf{m''})=
    \left\lVert 
h(\mathbf{m'}) -
h(\mathbf{m''})
    \right\rVert
    $ is an Euclidean pseudometric on $\mathcal{M}$ to $\mathbb{R}^d$. Since the set $\mathcal{Y}$ of all Euclidean pseudometrics from $\mathcal{M}$ to $\mathbb{R}^d$ is closed and complete, $\sigma((\boldsymbol{\Delta}^{(\infty)},\mathcal{P}),h)$ can be minimized by varying $h$ over the set of all Borel-measurable embedding functions from $\mathcal{M} \to \mathbb{R}^d$.
\end{remark}

%% file: text/setting.tex
\section{Description of the setting}
\label{Sec:Description_of_setting}
 A generative model is a random map from an input space or query space to an output space.
Let $f_1,\dots f_n$ be $ n $ generative models with the shared query space $\mathcal{Q}$ and the shared output space $\mathcal{X}$, and suppose $\{ q_1,\dots q_m\}$ is a collection of $ m $ queries.
In our setting, every model $f_i$ responds to every query $q_j$ exactly $r$ times and we let $f_i(q_j)_k$ denote the $k$-th replicate of the response of $f_i$ to $q_j$.
We assume that there exists an embedding function $g$ that maps a response to a vector in $\mathbb{R}^s$. 
 We let $ F_{ij} $ denote the probability distribution of the vector-embedded responses
 of $f_i$ to $q_j$, that is,
  $ g(f_i(q_j)_k) \sim^{iid} F_{ij} $ for all $i,j,k$.

For every $i$, the model $f_i$ is represented by the matrix $ \bar{\mathbf{X}}_{i} \in \mathbb{R}^{m \times s}$   whose $ j^{th} $ row is the mean over replicates of the $ i^{th} $ model's response to the $ j^{th} $ query, that is: $ (\bar{\mathbf{X}_{i}})_{j\cdot} = \frac{1}{r} \sum_{k=1}^{r} g(f_{i}(q_{j})_{k}) \approx (\boldsymbol{\mu}_{i})_{j\cdot} \equiv \mathbb{E}_{F_{ij}}\left[g(f_{i}(q_{j})\right] $. 

To capture the sample pairwise differences between the models, we define the matrix $ \mathbf{D} \in \mathbb{R}^{n \times n} $  with entries $ \mathbf{D}_{ii'} = \frac{1}{m} || \bar{\mathbf{X}}_{i} - \bar{\mathbf{X}}_{i'} ||_{F} $. Note that, for fixed $n$ and $m$, as $ r \to \infty $, $ \mathbf{D}_{ii'} $ converges to the quantity $ \boldsymbol{\Delta}_{ii'} = \frac{1}{m} || \boldsymbol{\mu}_{i} - \boldsymbol{\mu}_{i'}||_{F} $. The matrix $\boldsymbol{\Delta}=\left( 
\boldsymbol{\Delta}_{i i'}
\right)_{i,i'=1}^n$
is known as the model mean discrepancy matrix.

Multidimensional scaling by raw stress minimization yields  $ \widehat{\boldsymbol{\psi}} = \mathrm{mds}(\mathbf{D}) \in \mathbb{R}^{n \times d}$, representing our collection of models as a configuration of $ n $ points in finite-dimensional Euclidean space $ \mathbb{R}^{d} $. 
The geometry of $\widehat{\boldsymbol{\psi}}$ approximates the model discrepancy geometry configuration matrix $\boldsymbol{\psi} = \mathrm{mds}(\boldsymbol{\Delta}) \in \mathbb{R}^{n \times d} $. 
We refer to the $i$-th row
$\boldsymbol{\psi}_i$ (respectively $\widehat{\boldsymbol{\psi}}_i$) of the matrix $\boldsymbol{\psi}$ (respectively $\widehat{\boldsymbol{\psi}}$)   
  as (our estimate of) the \textit{data kernel perspective space} (DKPS) representation of model $ f_{i} $ with respect to the set of queries $ \{q_{j}\} $.
The remainder of this paper details the consistency of the estimated DKPS as $ n, m $ and/or $ r $ grow.

\begin{remark}
    In every setting we discuss, $r \to \infty$. For quantities whose definition depends on $r$, such as $\widehat{\boldsymbol{\psi}}$, we sometimes emphasize the dependence by writing $\widehat{\boldsymbol{\psi}}^{(r)}$.
\end{remark}

%% file: text/consistency_aistats.tex
\section{Consistency of estimated model embeddings}
\label{Sec:Consistency_of_model_embeddings}

The consistency results herein are presented in progressively more complicated settings where the number of queries and/or the number of models grow with a growing number of replicates from each $ F_{ij} $.

\subsection{Fixed set of models and fixed set of queries}
\label{Subsec:fixed_m_fixed_n_growing_r}

We start with the simplest setting where the number of models and the number of queries both remain fixed and the number of replicates increases. 
In this setting, we can establish the consistency of $\widehat{\boldsymbol{\psi}}$ (up to an affine transformation) by direct use of \textit{Theorem 2} from \cite{trosset2024continuous}, which establishes the convergence guarantees for raw stress embeddings of a sequence of dissimilarity matrices of a fixed size approaching another dissimilarity matrix. 

\begin{theorem}
    \label{Th:r_perspective_consistency}
    Let $\widehat{\boldsymbol{\psi}} \in
    \mathrm{MDS}(\mathbf{D}) \subset \mathbb{R}^{n \times d}$. Then
    there exists a subsequence $\lbrace r_u \rbrace_{u=1}^{\infty}$ of $\lbrace r \rbrace_{r=1}^{\infty}$ such that
    for all pairs $(i,i') \in [n]^2$,
    \begin{equation*}
    \left(
      \left\lVert 
\widehat{\boldsymbol{\psi}}_i^{(r_u)}-
\widehat{\boldsymbol{\psi}}_{i'}^{(r_u)}
      \right\rVert
      -
      \left\lVert 
\boldsymbol{\psi}_i-
\boldsymbol{\psi}_{i'}
      \right\rVert
      \right) \to^P 0
      \text{ as $u \to \infty$},
    \end{equation*}
     where $\boldsymbol{\psi}=
     [\boldsymbol{\psi}_1|\boldsymbol{\psi}_2|\dots | \boldsymbol{\psi}_n]^T
     \in
    \mathrm{MDS}(\boldsymbol{\Delta}) \subset \mathbb{R}^{n \times d}
    $.
\end{theorem}

\textit{Theorem} \ref{Th:r_perspective_consistency} states that by prompting each model with each query with enough replicates $ r $, the perspective space obtained via the raw stress criterion is close to the true-but-unknown perspective space for the models with respect to $ \{q_{j}\} $.



\subsection{Fixed set of models and growing set of queries}
\label{Subsec:growing_queries_fixed_models}
We next address the consistency of $\widehat{\boldsymbol{\psi}}$ in settings where the number of queries grows but the collection of models is fixed. 
That is, $n$ remains fixed but $m \to \infty$ as $r \to \infty$.
Our results are adapted from the results from \cite{trosset2024continuous}, where they show that if a sequence of dissimilarity matrices of a fixed size converges, then the 
raw stress embeddings of each term in the sequence approaches the raw stress embeddings of the limiting dissimilarity matrix. 
While we could directly use this result to argue that
$\widehat{\boldsymbol{\psi}}$ approaches $\boldsymbol{\psi}$
for fixed $ n $ and $ m $ where $\mathbf{D}_{i i'} \to \boldsymbol{\Delta}_{i i'}$ in Theorem \ref{Th:r_perspective_consistency}, we cannot do so in the context of growing $ m $. 
In particular, as $ m $ increases, so too does the dimensionality of $\boldsymbol{\Delta}$. 
Hence, in order to satisfy the condition that $\mathbf{D}$ approaches a specific dissimilarity matrix as $ m, r \to \infty$, we need an additional assumption:




\begin{assumption}
\label{Asm:Asymptotic_Euclidean_realizability_fixed_n}
For some $q \in \mathbb{N}$,
    there exist vectors $\boldsymbol{\phi}_1,\dots \boldsymbol{\phi}_n \in  \mathbb{R}^q$,  such that for every pair $(i,i') \in [n]^2$, 
    $
    \boldsymbol{\Delta}_{i i'} = \frac{1}{m}||\boldsymbol{\mu}_{i} - \boldsymbol{\mu}_{i'} ||
    \to
    \left\lVert 
\boldsymbol{\phi}_i-
\boldsymbol{\phi}_{i'}
    \right\rVert
    $ as $r, m \to \infty$.
\end{assumption}

Assumption \ref{Asm:Asymptotic_Euclidean_realizability_fixed_n} presumes the existence of high-dimensional vector representations for each generative model.
Our results will show that the sequence of low-dimensional raw stress configurations converges to a low-dimensional approximation of the $ \{\phi_{i}\}$ under appropriate conditions.
Let $\boldsymbol{\Delta}^{(\infty)}=
\left\lbrace 
\left\lVert 
\boldsymbol{\phi}_i
-
\boldsymbol{\phi}_{i'}
\right\rVert
\right\rbrace_{i,i'=1}^{n} \in \mathbb{R}^{n \times n}
$ and $\boldsymbol{\psi}=\mathrm{mds}(\boldsymbol{\Delta}^{(\infty)}) \in 
\mathrm{MDS}(\boldsymbol{\Delta}^{(\infty)}) \subset
\mathbb{R}^{n \times d}$.
Note that the vectors $\boldsymbol{\phi}_i$ being independent of $m$ and $r$ ensures that the dissimilarity matrices $\mathbf{D}$ approach a specific limit $\boldsymbol{\Delta}^{(\infty)}$ as $m, r \to \infty$.
Using results from \cite{trosset2024continuous}, we show that if $\mathbf{D}$ approaches the model mean discrepancy matrix $\boldsymbol{\Delta}^{(\infty)}$ then  $\widehat{\boldsymbol{\psi}}$ approaches 
$\boldsymbol{\psi}$.

\begin{lemma}
\label{Lm:RS_minimizer_convergence_fixed_n}
(\cite{trosset2024continuous}) 
Let $ \widehat{\boldsymbol{\psi}} \in
\mathrm{MDS}(\mathbf{D}) \subset \mathbb{R}^{n \times d}$.
Let $ n $ be fixed and let $ m $ grow with $ r $.
If $\left\lVert \mathbf{D}- \boldsymbol{\Delta}^{(\infty)} \right\rVert_F \to^P 0$ as $r \to \infty$, then there exists a subsequence $\lbrace r_u \rbrace_{u=1}^{\infty}$ of $\lbrace r \rbrace_{r=1}^{\infty}$ such that
for all pairs $(i,i') \in [n]^2$,
\begin{equation*}
\left(
    \left\lVert 
\widehat{\boldsymbol{\psi}}_i^{(r_u)}-
\widehat{\boldsymbol{\psi}}_{i'}^{(r_u)}
\right\rVert -
\left\lVert 
\boldsymbol{\psi}_i -
\boldsymbol{\psi}_{i'}
\right\rVert
\right) \to^P 0
\text{ as $u \to \infty$},
\end{equation*}
 where $\boldsymbol{\psi}=
 [\boldsymbol{\psi}_1|\boldsymbol{\psi}_2|\dots |\boldsymbol{\psi}_n]^T
 \in \mathrm{MDS}(\boldsymbol{\Delta}^{(\infty)}) \subset \mathbb{R}^{n \times d}$.
\end{lemma}


We note that by the weak law of large numbers we have $
(\bar{\mathbf{X}}_i)_{j \cdot} \to^P
(\boldsymbol{\mu}_i)_{j \cdot}
$ as $ r \to \infty $.
However, if $ m $ is large while $ r $ is relatively small, then there is non-zero probability that many rows in $ \bar{\mathbf{X}}_{i} $ are far away from the corresponding rows in $ \boldsymbol{\mu_{i}
}$.
Thus, if $ m $ grows too fast with respect to $ r $, $ \mathbf{D} $ may not approach $ \boldsymbol{\Delta}^{(\infty)} $ and the condition of \textit{Lemma \ref{Lm:RS_minimizer_convergence_fixed_n}} will not be met.
We next establish a sufficient condition for the rate at which $ m $ can grow with respect to $ r $ for the condition of \textit{Lemma \ref{Lm:RS_minimizer_convergence_fixed_n}}.


\begin{theorem}
\label{Th:sufficient_condition}
Let $ \boldsymbol{\Sigma}_{ij}=\mathrm{cov}[g(f_i(q_j)_k)]$ be the covariance matrix associated with the distribution $F_{ij}$ and $\gamma_{ij}=\mathrm{trace}(\boldsymbol{\Sigma}_{ij})$. 
If, for all $i \in [n]$,
\begin{equation*}
   \lim_{r \to \infty}
    \frac{ \frac{1}{m}
    \sum_{j=1}^{m} \gamma_{ij}}{r}=0
\end{equation*}
then
     $
     \left\lVert
\mathbf{D}-\boldsymbol{\Delta}^{(\infty)}  
\right\rVert_F \to^P 0
     $ as $r \to \infty$.
\end{theorem}


As a consequence of \textit{Lemma \ref{Lm:RS_minimizer_convergence_fixed_n}} and \textit{Theorem \ref{Th:sufficient_condition}}, a sufficient condition for consistency of $\widehat{\boldsymbol{\psi}}$ is
$\frac{1}{m} \sum_{j=1}^m \gamma_{ij}=o(r)$ for all $i \in [n]$. 
\begin{theorem}
\label{Th:main}
    In the setting of \textit{Theorem \ref{Th:sufficient_condition}},
    suppose for all $i \in [n]$, $\frac{1}{m} \sum_{j=1}^{m} \gamma_{ij}=
o(r)$.
Also, let $\widehat{\boldsymbol{\psi}} \in 
\mathrm{MDS}(\mathbf{D}) \subset \mathbb{R}^{n \times d}$.
There exists a subsequence $\lbrace r_u \rbrace_{u=1}^{\infty}$ of $\lbrace r \rbrace_{r=1}^{\infty}$ such that
 for all $(i,i') \in [n]^2$, 
 \begin{equation*}
 \left(
     \left\lVert 
\widehat{\boldsymbol{\psi}}_i^{(r_u)}-
\widehat{\boldsymbol{\psi}}_{i'}^{(r_u)}
\right\rVert -
\left\lVert 
\boldsymbol{\psi}_i
-
\boldsymbol{\psi}_{i'}
\right\rVert
\right) \to^P 0
\text{ as $u \to \infty$},
 \end{equation*}
  where $\boldsymbol{\psi}=
  [\boldsymbol{\psi}_1|\boldsymbol{\psi}_2|\dots |\boldsymbol{\psi}_n]^T
  \in \mathrm{MDS}(\boldsymbol{\Delta}^{(\infty)}) \subset \mathbb{R}^{n \times d}$.
\end{theorem}
Thus, \textit{Theorem \ref{Th:main}} gives us an idea of how fast $ m $ can grow with respect to $ r $ as a function of $ F_{ij}$, in order for $\widehat{\boldsymbol{\psi}}$ to be consistent. 
Observe that if $\max_{j \in [m]}\gamma_{ij}=O(1)$ for all $i \in [n]$, then $ m $ can grow arbitrarily fast with respect to $ r $, and $\widehat{\boldsymbol{\psi}}$ will still be consistent. 

As a consequence of \textit{Theorem \ref{Th:main}}, each $\widehat{\boldsymbol{\psi}}_i$ will accumulate to some affine transformation of $\boldsymbol{\psi}_i$, which is stated in the following corollary.
\begin{corollary}
\label{cor:perspectives}
    In the setting of \textit{Theorem \ref{Th:main}},
    suppose for all $i \in [n]$, $\frac{1}{m} \sum_{j=1}^m \gamma_{ij}=o(r)$. Then
    there exist sequences 
    $\left\lbrace
 \mathbf{W}^{(u)} \right\rbrace_{u=1}^{\infty}$ and 
 $\left\lbrace
 \mathbf{a}^{(u)}
 \right\rbrace_{u=1}^{\infty}$,
 satisfying $\mathbf{W}^{(u)} \in \mathcal{O}(d)$ and $\mathbf{a}^{(u)} \in \mathbb{R}^d$ for all $u \in \mathbb{N}$,
 such that for all $i \in [n]$,
    \begin{equation*}
        \left\lVert 
\widehat{\boldsymbol{\psi}}_i^{(r_u)}
-
\left(
\mathbf{W}^{(u)}
\boldsymbol{\psi}_i
+
\mathbf{a}^{(u)}
\right)
        \right\rVert
        \to^P 0
        \text{ as $u \to \infty$}.
    \end{equation*}
\end{corollary}
Corollary \ref{cor:perspectives} states that there is a subsequence of minimizers of the raw stress criterion that is close to the true-but-unknown perspectives as the number of queries and number of replicates for each query grow.
In our set up the true-but-unknown perspective space depends on the collection of models and the growing set of queries.
While our results allow for arbitrary growing sets of queries, the true-but-unknown perspective space is most easily characterized when the queries come from an explicit query distribution.

\subsection{Growing set of models and growing set of queries}
\label{Subsec:growing_queries_growing_models}

We now let both the number of models and the number of queries grow as the number of replicates grow 
and establish that under appropriate conditions $\widehat{\boldsymbol{\psi}}$ is consistent.
In this setting, we need an adaptation of \textit{Assumption} 1 that further assumes that the vectors that induce the limiting dissimilarity are elements of a compact Riemannian manifold:
\begin{assumption}
    \label{Asm:Asymptotic_Euclidean_realizability_growing_n}
    Let $ \mathcal{M} $ be a compact Riemannian manifold.
    For every model $f_i$, there exists a vector $\boldsymbol{\phi}_i \in \mathcal{M} \subset \mathbb{R}^q$, such that for all pairs $(i,i') \in \mathbb{N} \times \mathbb{N}$,
    $\frac{1}{m}
    \left\lVert 
    \boldsymbol{\mu}_i-
    \boldsymbol{\mu}_{i'}
    \right\rVert
    \to 
    \left\lVert 
\boldsymbol{\phi}_i-
\boldsymbol{\phi}_{i'}
    \right\rVert
    $ as $r \to \infty$. 
\end{assumption}
Recall that in the growing sample size scenario, the definition of raw stress relies on a dissimilarity function. 
In our setting, we define 
the following
dissimilarity function $\boldsymbol{\Delta}^{(\infty)}:\mathcal{M} \times \mathcal{M} \to \mathbb{R}_{\ge 0}$ to be $\boldsymbol{\Delta}^{(\infty)}(\boldsymbol{\phi}_i,\boldsymbol{\phi}_{i'})=
\left\lVert 
\boldsymbol{\phi}_i-
\boldsymbol{\phi}_{i'}
\right\rVert
$. 
Results from \cite{trosset2024continuous} tell us that $\widehat{\boldsymbol{\psi}}$ is consistent if $\mathbf{D}_{i i'}$ approaches 
$
\boldsymbol{\Delta}^{(\infty)}
(\boldsymbol{\phi}_i,\boldsymbol{\phi}_{i'})
$ for all pairs $i,i' \in [n] $.

\begin{lemma}
(\cite{trosset2024continuous})
\label{Lm:D_to_psi_consistency_growing_n}
    Let $\boldsymbol{\phi}_i \sim^{iid} \mathcal{P}$. Assume that for all pairs $(i,i') \in \mathbb{N} \times \mathbb{N}$, 
    $\mathbf{D}_{i i'} \to^P \boldsymbol{\Delta}^{(\infty)}(\boldsymbol{\phi}_i,\boldsymbol{\phi}_{i'})$ as $r \to \infty$.
    Then, for some subsequence $\lbrace r_u \rbrace_{u=1}^{\infty}$ of $\lbrace r \rbrace_{r=1}^{\infty}$, for all $p \geq 1$,
\begin{align*}
&\int_{\mathcal{M}}
\int_{\mathcal{M}}
\left(
\left\lVert 
\widehat{\boldsymbol{\psi}}_1^{(r_u)}-
\widehat{\boldsymbol{\psi}}_2^{(r_u)}
\right\rVert -
\left\lVert 
\mathrm{mds}(\boldsymbol{\phi}_1)-
\mathrm{mds}(\boldsymbol{\phi}_2)
\right\rVert
\right)^p \\
\nonumber & \quad \quad \quad \quad \quad \quad \quad \quad \quad \mathcal{P}(d \boldsymbol{\phi}_1)
\mathcal{P}(d \boldsymbol{\phi}_2)
\to^P
0 \text{ as $u \to \infty$}.
\end{align*}
\end{lemma}

\begin{figure*}[t!]
    \centering
    \begin{subfigure}{0.49\textwidth}
        \centering
    \includegraphics[width=\textwidth]{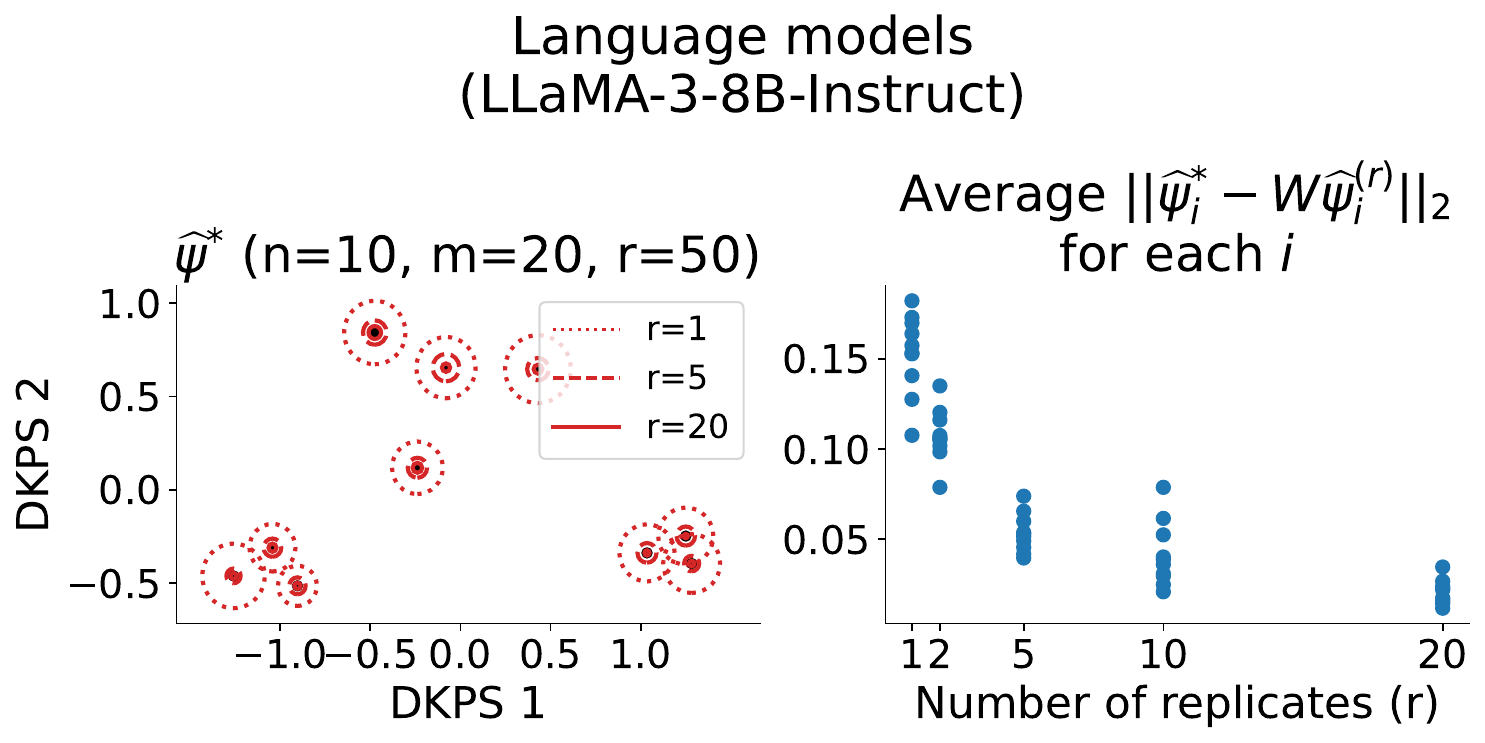}
    \end{subfigure}
    \begin{subfigure}{0.49\textwidth}
        \centering
    \includegraphics[width=\textwidth]{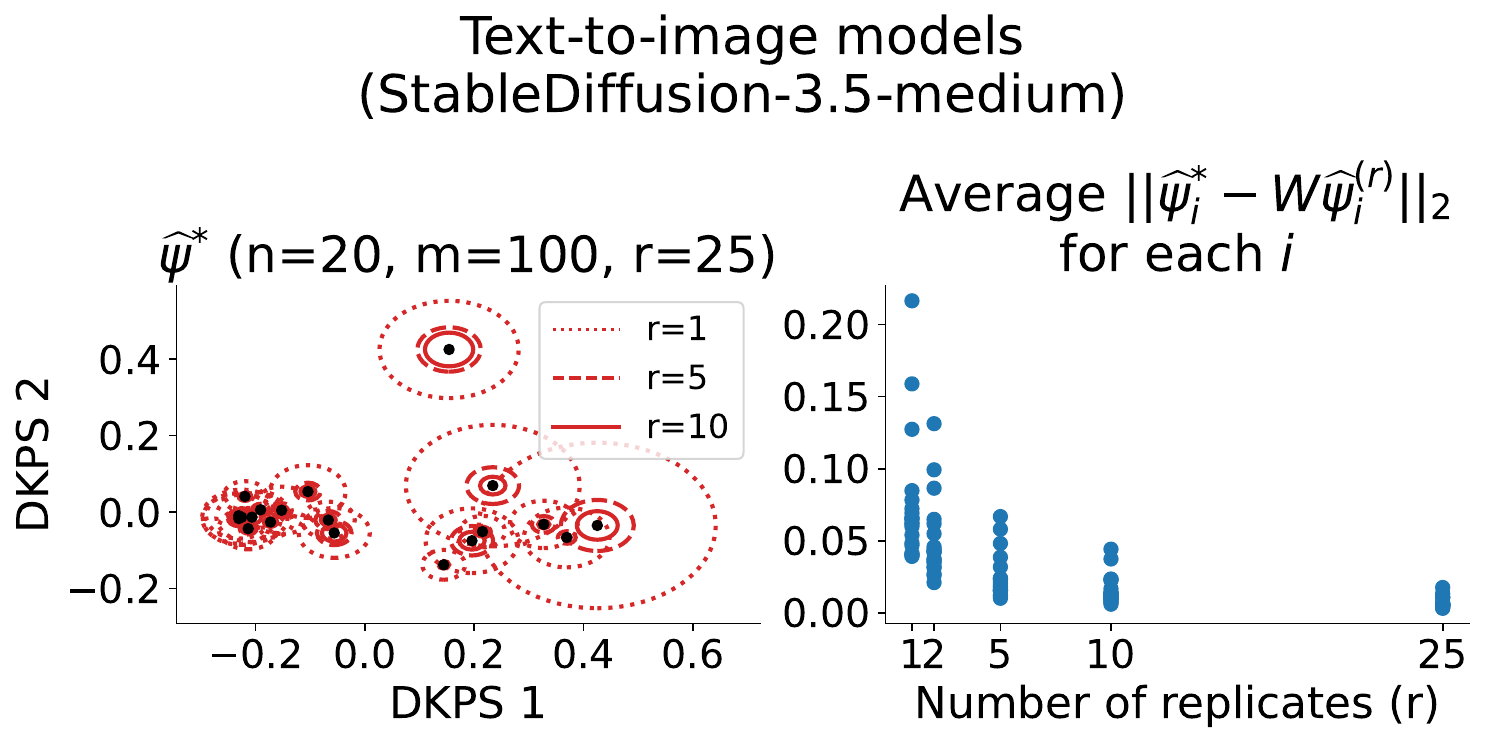}
    \end{subfigure}
    \caption{Numerical evidence of the consistency of $ \widehat{\boldsymbol{\psi}} $ to $ \boldsymbol{\psi} $ for fixed $ n $ and fixed $ m $ for a collection of language models (left) and a collection of text-to-image models (right). 
    The black dots in the left figure of each pair are the 2-d perspectives of models induced with randomly selected queries, $ R $ replicates each, and a domain-specific embedding function. 
    The red circles have radius equal to the average $ L_{2} $ between $ \widehat{\psi}^{*} $ and model representations estimated with $ r $ replicates for each query. 
    The right figure of each pair shows the distribution of the average $ L_{2} $ norm in the DKPS across models for various $ r $.
    More replicates improves estimation quality.}
    \label{fig:increasing_r}
\end{figure*}

The above theorem establishes that if the pairwise distances  $\mathbf{D}_{i i'}$  approach the dissimilarities $\boldsymbol{\Delta}^{(\infty)}(\boldsymbol{\phi}_i,
\boldsymbol{\phi}_{i'})$, then $\hat{\boldsymbol{\psi}}$ is consistent. In order for 
$\mathbf{D}_{i i'}$ to approach $\boldsymbol{\Delta}^{(\infty)}(\boldsymbol{\phi}_i,
\boldsymbol{\phi}_{i'})$, a result similar to \textit{Theorem \ref{Th:sufficient_condition}} is needed, which holds in the regime of growing $n$. Such a result can be shown to hold true by the same equipment used to establish \textit{Theorem \ref{Th:sufficient_condition}}.
\begin{theorem}
\label{Th:sufficient_condition_growing_n}
In our setting where $m,n \to \infty$ as $r \to \infty$, $|\mathbf{D}_{i i'}-\boldsymbol{\Delta}^{(\infty)}
(\boldsymbol{\phi}_i,
\boldsymbol{\phi}_{i'})
| \to^P 0$ for every pair $(i,i') \in \mathbb{N} \times \mathbb{N}$, if 
\begin{equation*}
    \lim_{r \to \infty}
    \frac
    {
    \frac{1}{m} \sum_{j=1}^m \gamma_{ij}
    }
    {
    r
    } =0
    \hspace{0.2cm} 
    \text{for all $i$}.
\end{equation*}
\end{theorem}
 It may seem a bit surprising at first that the sufficient conditions in both scenarios are the same: The setting with growing $ m $ and $ n $ was expected to be more restrictive. 
 Note that in both cases (of fixed $n$ and growing $n$), it is pointwise convergence of $\mathbf{D}_{i i'}$ to $\boldsymbol{\Delta}^{(\infty)}(\boldsymbol{\phi}_i,
 \boldsymbol{\phi}_{i'})$ that we need. That is why in each of the cases it suffices to ensure that $\frac{1}{m} \sum_{j=1}^m \gamma_{ij}=o(r)$ for all $i$. Further, the condition that for all $i$ 
$\frac{1}{m} \sum_{j=1}^m \gamma_{ij}= o(r)$ is more difficult to ensure when $n$ grows than when $n$
is fixed. 

As in \textit{Section \ref{Subsec:growing_queries_fixed_models}},
the lemmas in this section allow us to deduce a sufficient condition for the consistency of 
$\widehat{\boldsymbol{\psi}}$:
\begin{theorem}
    \label{Th:main_growing_n}
    In the setting of \textit{Lemma \ref{Lm:D_to_psi_consistency_growing_n}}, 
    suppose for all $i \in \mathbb{N}$, 
    $\frac{1}{m} \sum_{j=1}^m \gamma_{ij}=o(r)$. Then,
    for all $p \geq 1$, for some subsequence $\lbrace r_u \rbrace_{u=1}^{\infty}$ of $\lbrace r \rbrace_{r=1}^{\infty}$,
\begin{align*}
 \int_{\mathcal{M}}
\int_{\mathcal{M}} 
        \left(
        \left\lVert 
\widehat{\boldsymbol{\psi}}_1^{(r_u)}-
\widehat{\boldsymbol{\psi}}_2^{(r_u)}
        \right\rVert -
        \left\lVert 
\mathrm{mds}(\boldsymbol{\phi}_1) -
\mathrm{mds}(\boldsymbol{\phi}_2)
        \right\rVert
        \right)^p 
         \mathcal{P}(d \boldsymbol{\phi}_1)
        \mathcal{P} (d \boldsymbol{\phi}_2)
        \to^P
        0
        \text{ as $u \to \infty$}.
    \end{align*}
\end{theorem}
Thus, even in the case where $n$ grows, $m$ is allowed to grow arbitrarily fast with respect to $r$ if $\max_{j \in [m]} \gamma_{ij}=O(1)$ for all $i$. Moreover, since all we need is pointwise convergence of $\mathbf{D}$ to $\boldsymbol{\Delta}^{(\infty)}$, $n$ can grow arbitrarily fast with respect to $r$.

%% file: text/numerical_aistats.tex
\section{Numerical experiments}
\label{Sec:Numerical_exp}

We next provide empirical support of the consistency of the representations of models in the three settings we analyzed above.
For each setting, we study a collection of large language models and a collection of text-to image models.

\subsubsection*{Language models} For our language model example, we study collections of different \texttt{LLaMA-3-8B-Chat} \citep{dubey2024llama} models. 
The models are parameterized by different fixed context augmentations $ a_{i} $, i.e., $ f_{i} = f(\;\cdot\;; a_{i}) $. 
Each $ a_{i} $ is a text string related to RA Fisher such as $ a_{i} =$ ``RA Fisher pioneered the principles of the design of experiments.” or $ a_{i'} =$ ``RA Fisher’s view on eugenics were primarily based on anecdotes and prejudice." written by us. 
Given a query $ q_{j} $, the base model is prompted with the appropriately formatted prompt $ ``a_{i} \; q_{j}" $ $ R=50 $ times. 
In our experiments we consider up to $ 174 $ questions about RA Fisher such as $ q_{j} =$ ``What is R.A. Fisher's most well-known statistical theorem?"
as queries.
We consider up to $ N=50 $ models.
The questions were sampled from ChatGPT with the prompt ``Provide 200 questions related to RA Fisher". 
We did not include a random $ 26 $ questions.
We use the open source embedding model \texttt{nomic-embed-text-v1.5} \citep{nussbaum2024nomicembedtrainingreproducible} to construct $ \bar{\mathbf{X}}_{i} \in \mathbb{R}^{m \times 756} $ and GrasPy's \citep{JMLR:v20:19-490} implementation of multi-dimensional scaling to map the collection of $ \bar{\mathbf{X}}_{i} $ to $ \mathbb{R}^{2}$.

\begin{figure*}[t]
\centering
    \begin{subfigure}{0.49\textwidth}
        \centering
    \includegraphics[width=\textwidth]{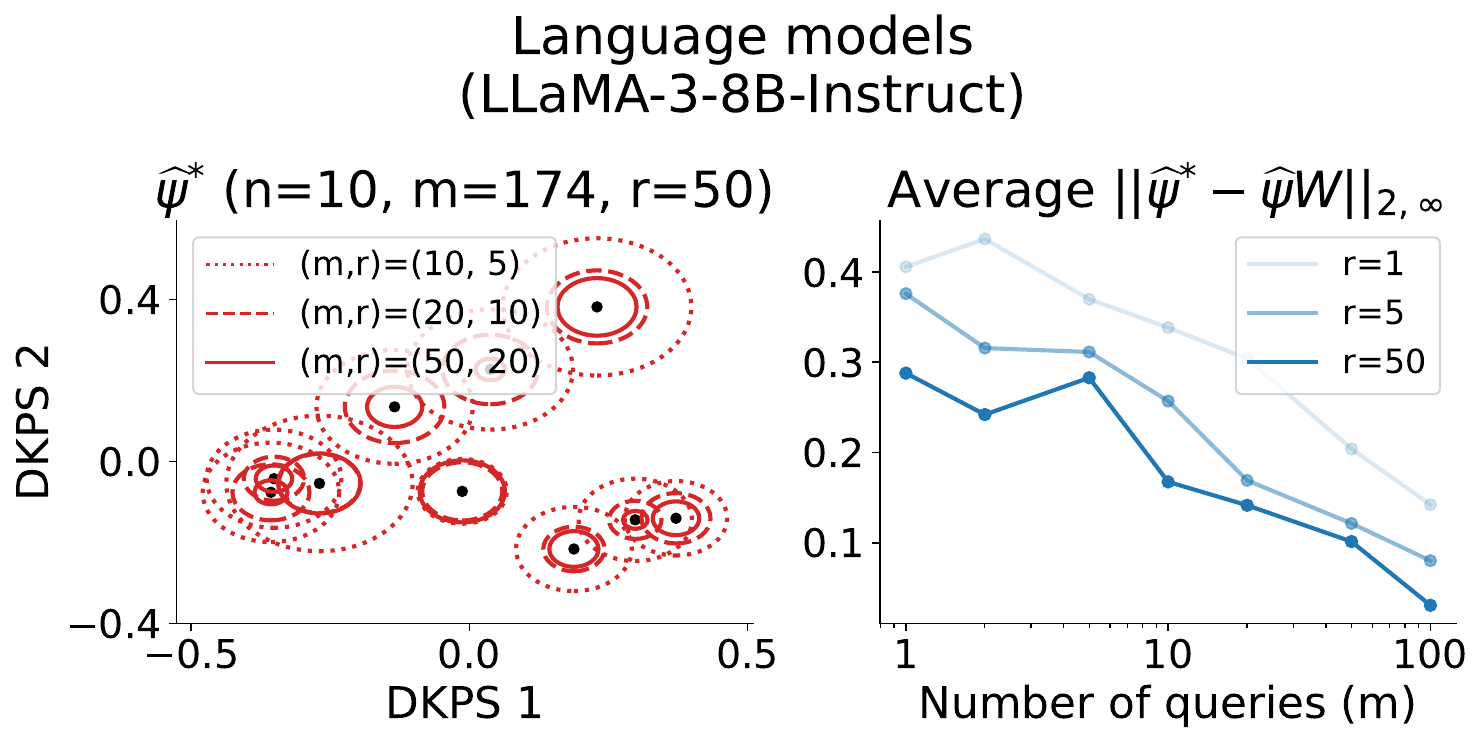}
    \end{subfigure}
    \begin{subfigure}{0.49\textwidth}
        \centering
    \includegraphics[width=\textwidth]{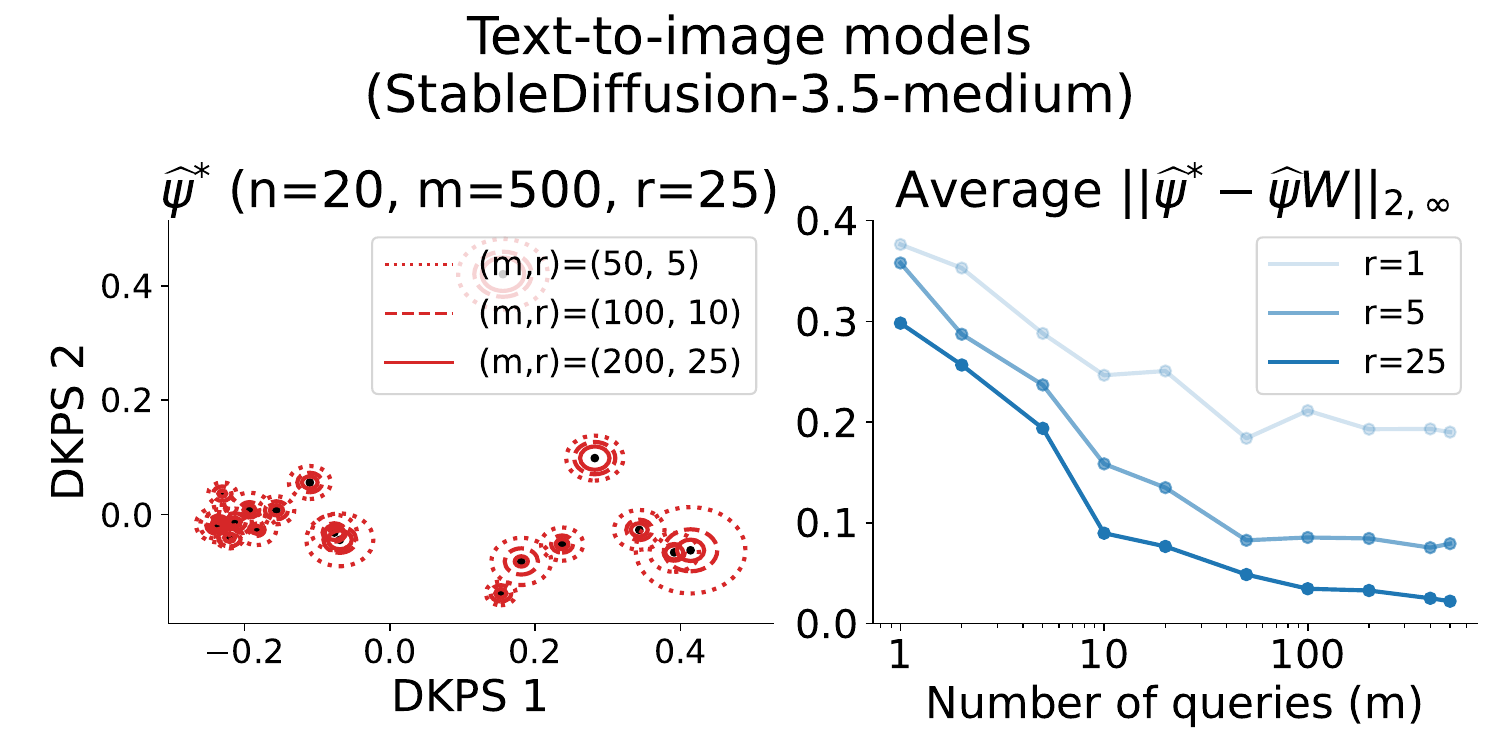}
    \end{subfigure}
    \caption{Numerical evidence of the consistency of $ \widehat{\boldsymbol{\psi}} $ to $ \boldsymbol{\psi} $ for fixed $ n $ and growing $ m $. 
    The black dots in the left of each pair of figures are the 2-d perspectives of $ n $ models induced with $ M $ queries and $ R $ replicates each. The red circles have radius equal to the average $ L_{2} $ norm between the ``ground truth" and model representations estimated for selected $ (m, r) $ pairs. 
    The right figures show the average maximum row $L_{2}$ norm for various $ (m, r) $ pairs. 
    More replicates and more queries improves estimation quality. 
    The number of queries appears to have a larger effect.}
    \label{fig:increasing_m_r}
\end{figure*}

\subsubsection*{Text-to-image models} For our text-to-image example, we study collections of \texttt{StableDiffusion-3.5-medium} \citep{esser2024scaling}. 
As with our language model example, we parameterize different models by different fixed context augmentations.
Each fixed augmentation $ a_{i} $ is an instruction to produce an image in the style of a famous artist -- 
for example, $ a_{i} $ = ``in the style of Pablo Picasso" or $ a_{i'} $ = ``in the style of Leonardo da Vinci" -- and $ f_{i} = f(\; \cdot\;; a_{i}) $. 
We consider up to $ N = 100 $ models.
Given a query $ q_{j} $, the base model is prompted with the appropriately formatted prompt $ ``q_{j} \; a_{i}  " $ $ R=25 $ times. 
Each query $ q_{j} $ is an instruction to produce an image of a noun.
For example, one of the prompts for the model corresponding to da Vinci is ``An apple in the style of Leonardo da Vinci".
The nouns were generated from ChatGPT with the prompt ``Provide 500 nouns."
We use the open source embedding model \texttt{nomic-embed-vision-v1.5} to map the generated images to a vector space and construct the $ \bar{\mathbf{X}}_{i} \in \mathbb{R}^{m \times 768} $.
We again use GrasPy's implementation of multi-dimensional scaling to induce the data kernel perspective space of the models.
To avoid sounding repetitive, we use $ N $, $ M $, and $ R $ to mean the maximum number of models, queries, and replicates for both the language model experiments and the text-to-image experiments.
For the language model experiment, $ N = 50 $, $ M = 174 $, and $ R = 50 $. For the text-to-image experiment, $ N = 100$, $ M = 500 $, and $ R = 25 $.

For both sets of experiments we do not have the ``ground truth" DKPS $ \boldsymbol{\psi} $ in any parameter setting, and thus treat $ \widehat{\boldsymbol{\psi}} $ estimated with $ r= R $ as a proxy for the ground truth.
For settings with growing $ n $ and/or growing $ m $, we similarly treat $ \widehat{\boldsymbol{\psi}}^{(R)} $ estimated with $ n = N $ and $ m = M $ as the ground truth.
We refer to these estimates as $ \widehat{\boldsymbol{\psi}}^{*} $.
We let $ d = 2 $ in all settings for visualization purposes.
Finally, in settings where $ n, m, $ or $ r $ grow, we sample from the appropriate set of models, queries, or replicates with replacement to calculate $ \widehat{\boldsymbol{\psi}} $ and report the average $ L_{2} $ norm or average two-to-infinity norm of the difference between $ \boldsymbol{\widehat{\psi}}^{*} $ and $ \boldsymbol{\widehat{\psi}} \mathbf{W} $ of 10 bootstrap samples, where $ \mathbf{W} \in \mathcal{O}(2) $ is the Procrustes solution \citep{goodall1991procrustes}.

\subsection{Fixed collection of models and fixed set of queries}
As with the theoretical analysis, our empirical analysis starts in the simplest setting where $ n $ and $ m $ are fixed while $ r $ grows. 
The collection of models and set of queries were selected at random without replacement -- $ n = 10 $ and $ m = 20 $ for the language model example, $ n = 20 $ and $ m = 100 $ for the text-to-image example.
The target DKPS $ \boldsymbol{\widehat{\psi}}^{*} $ for the two experiments are shown on the left for each pair of figures of Figure \ref{fig:increasing_r}. 
Each black dot represents a model. 
The red circles have radius equal to the average (across bootstrap samples) $ L_{2} $ norm between $ \boldsymbol{\widehat{\psi}}^{*} $ and $ \boldsymbol{\widehat{\psi}}^{(r)} \mathbf{W} $ for selected $ r $. 
The right figure for each pair of figures of Figure \ref{fig:increasing_r} shows the distribution of the average (across bootstrap samples) $ L_{2} $ norm for each model for more values of $ r $.
The decreasing average $ L_{2} $ norm for all models as $ r $ increases supports Theorem \ref{Th:r_perspective_consistency}.

\subsection{Fixed collection of models and growing set of queries}
We next consider the setting where $ n $ is fixed and both $ m $ and $ r $ grow.
We use the same models as above. 
The two target DKPS $ \widehat{\boldsymbol{\psi}}^{*} $ are shown on the left for each pair of Figure $ \ref{fig:increasing_m_r} $. 
They were estimated with $ m = M $ and $ r = R $.
As with Figure \ref{fig:increasing_r}, the red circles have radius equal to the average (across bootstrap samples) $ L_{2} $ norm of the difference between $ \widehat{\boldsymbol{\psi}}^{*} $ and $ \widehat{\boldsymbol{\psi}} \mathbf{W} $ for selected $ (m, r) $ pairs.
The right figure of Figure \ref{fig:increasing_m_r} shows the average two-to-infinity norm of the difference between $ \widehat{\boldsymbol{\psi}}^{*} $ and $ \widehat{\boldsymbol{\psi}} $ across bootstrap samples for more values of $ (m, r) $. 
The two-to-infinity norm decreases as both $ (m, r) $ increase, which supports Theorem \ref{Th:main}.

The right figure of each pair figures of Figure \ref{fig:increasing_m_r} shows the possibility of a computational trade-off between increasing $ m $ and increasing $ r $. 
In particular, the computational cost of each iteration of the experiment is approximately $ O(m r) $: 
getting more replicates of a fixed number of queries is approximately the same as getting a small set of replicates for more queries from a compute stand point.
From an estimation stand point, however, the right figures of Figure \ref{fig:increasing_m_r} shows that getting more queries may be more beneficial.
Per Theorem \ref{Th:sufficient_condition}, this observation depends on the distributional properties of each $ F_{ij} $.

\subsection{Growing collection of models and growing set of queries}

\begin{figure*}[t]
\centering
    \begin{subfigure}{0.30\textwidth}
        \centering
    \includegraphics[width=\textwidth]{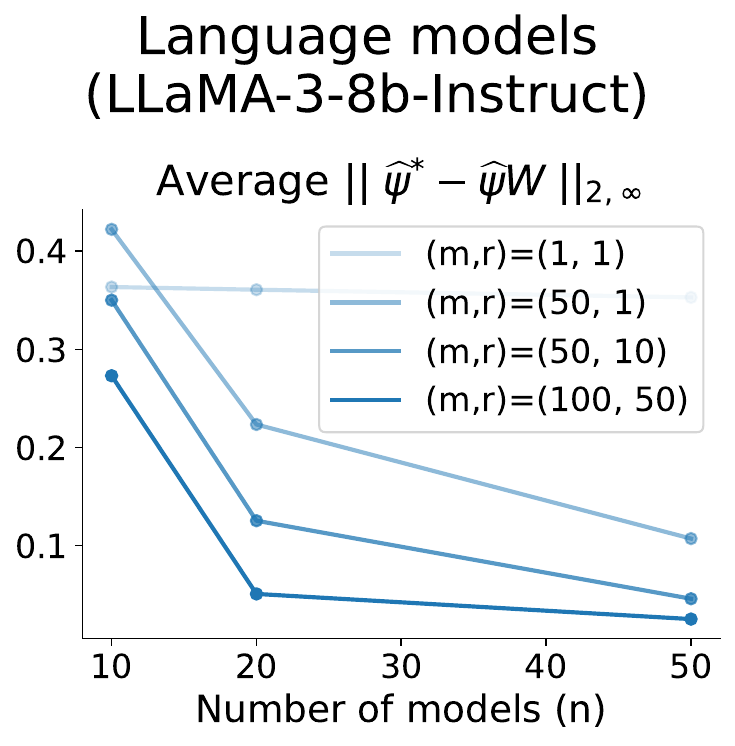}
    \end{subfigure}
    \begin{subfigure}{0.33\textwidth}
        \centering
    \includegraphics[width=\textwidth]{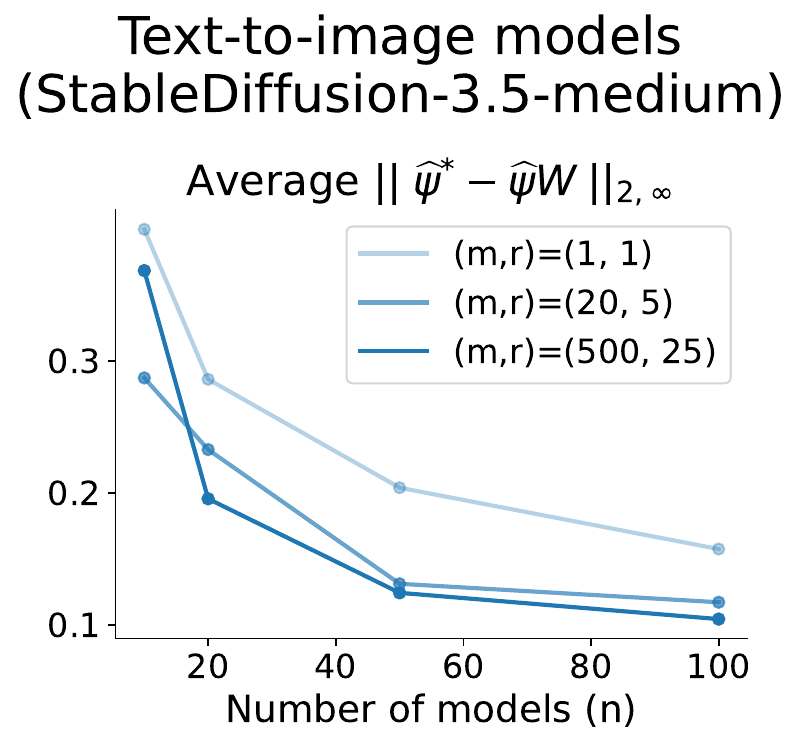}
    \end{subfigure}
    \caption{Numerical evidence of the consistency of $ \widehat{\boldsymbol{\psi}} $ to $ \boldsymbol{\psi} $ for growing $ n $ and growing $ m $. 
    The two target DKPS $ \widehat{\psi} $ were estimated using $ N $, $ M $, and $ R $.}
\label{fig:increasing_n_m_r}
\end{figure*}

Finally, we consider the setting with growing $ n $ and growing $ m $. 
The two target DKPS $ \widehat{\boldsymbol{\psi}}^{*} $ were estimated using all $ N $ models, all $ M $ queries and all $ R $ replicates per query. 
For each $ (n, m, r) $ triple, we randomly sample $ n $ models, $ m $ queries, and $ r $ replicates with replacement and estimate the DKPS $ \widehat{\boldsymbol{\psi}} $.
Figure \ref{fig:increasing_n_m_r} shows the average two-to-infinity norm of the difference between $ \widehat{\boldsymbol{\psi}}^{*} $ and $ \widehat{\boldsymbol{\psi}} \mathbf{W} $ across bootstrap samples for the $ n $ sampled models.
As each element of the triple increases, the reported difference gets close to $ 0 $, which supports Theorem \ref{Th:main_growing_n}.

Similar to the difference in the effect on estimation between $ r $ and $ m $, we observe a similar difference in the effect of estimation between $ r $ and $ n $ in Figure \ref{fig:increasing_n_m_r}.
In particular, increasing the number of models appears to have a larger effect on estimation than increasing $ r $ (or, even increasing $ m $).
As above, this observation depends on the $ F_{ij} $, per Theorem \ref{Th:sufficient_condition_growing_n}.
Unlike the observation in Figure \ref{fig:increasing_m_r}, however, the computational cost of increasing $ n $ is larger than increasing $ m $ or $ r $ since $ m \cdot r $ samples from $ F_{ij} $ are required for each additional $ n $.
Thus, there is a more delicate trade-off between computational cost and estimation quality in this setting than in the fixed $ n $ case.

%% file: text/discussion_aitstats.tex
\section{Discussion}
\label{Sec:Discussion}
In \cite{helm2024tracking}, a novel method was proposed for embedding generative models into a finite-dimensional Euclidean space in the context of queries -- the \textit{data kernel perspective space}.
In this paper, we analyzed the estimated perspective space in regimes where the number of models and/or the number of queries can remain fixed or grow and demonstrated that it is consistent for a population-level perspective space.
Importantly, we also describe sufficient conditions for consistent estimation of the perspective space. 
In this regard, we establish that if the number of queries and/or models grow adequately slowly with the number of replicates relative to distributional properties of the $ F_{ij} $ then the perspective space can be estimated consistently. 

Low-dimensional representations of collections of generative models enable the use of classical methods to understand differences in model behavior and, eventually, to make sense of model evolution.
For example, \cite{helm2024tracking} uses the perspective space to demonstrate that the communication structure underlying a system of interacting language models has an impact on system-level and model-level dynamics.
Going further, the perspective space can be used for model-level inference problems such as predicting the the pre-training data mixture, predicting model safety, predicting the model's score on a benchmark, etc. 
Deeper investigations into these applications are warranted and our results support the perspective space as a principled foundation for these pursuits. 

While the consistency of the estimated perspective space holds for all choices of the dimensionality of the raw stress embeddings, we note that the choice will impact practical properties of the estimate.
In particular, choosing large $ d $ may result in slower convergence.
Conversely, choosing small $ d $ may result in fast convergence but to a limiting set of vectors that poorly approximate the high-dimensional $ \lbrace \boldsymbol{\phi}_i \rbrace $. 

Further, the consistency of the estimated perspectives is the least we could ask for \citep{bickel2015mathematical}. 
We expect concentration inequalities and distributional results for the estimated perspectives to be important theoretical extensions to support applications in non-asymptotic regimes.
For example, establishing uniform convergence of the perspectives would enable use of the estimated perspective space as a principled substitute for its population counterpart after a particular amount of models, queries, and replicates. 
Similarly, establishing distributional properties of the estimated perspectives, such as asymptotic normality, in particular settings as to practicable inference methods. 

We required Assumptions 1 and 2 to establish consistency of $ \widehat{\boldsymbol{\psi}} $ in regimes for growing $ m $ with either fixed or growing $ n $.
In particular, we assume the existence of a collection of vectors in a finite and fixed $ q $-dimensional Euclidean space whose inner point distances are the limiting dissimilarities between the perspectives.
These two assumptions put implicit constraints on how the queries grow and on the properties of each $ F_{ij} $. 
An important and more general follow-on investigation may relax this assumption such that $ \lbrace \boldsymbol{\phi}_i \rbrace $ exist in a reproducing kernel Hilbert space.

We note that for most generative models in practice such as large language models with a maximum context size and maximum output sequence length and diffusion models that output RBG images, there are only a finite many possible input and output sequences or images. 
Hence, the maximum trace of the covariance matrices of $ \{F_{ij}\} $ does not grow without bound and, importantly, we are in the regime where $ m $ or $ n $ can grow arbitrarily fast with respect to $ r $.

Lastly, our results are for representations of models in the context of a set of queries.
There are other potential vector representations of models that do not depend on a set of queries, such as model weights.
The geometry of these quantities can similarly be obtained in settings where an appropriate dissimilarity is defined.
In our setting, different sets of queries will approximate the non-query dependent geometry to different degrees.
Investigations into how well different query sets approximate more inherent features of the models are necessary, as well as investigations into practical trade-offs between query-independent and query-dependent model representations. 